\documentclass{article}
\PassOptionsToPackage{numbers}{natbib}

\usepackage{amssymb, amsmath, amsthm, mymacros, graphicx, appendix} 
\usepackage{iclr2017_conference}
\iclrfinalcopy

\usepackage{float}
\usepackage{wrapfig}
\usepackage{booktabs}

\usepackage[stable]{footmisc}
\usepackage[numbers]{natbib}
\usepackage{url}
\usepackage{todonotes}
\usepackage{multirow}
\usepackage{pbox}
\usepackage{hhline}
\usepackage{subcaption}
\newcolumntype{L}{>{\centering\arraybackslash}m{2cm}}

\newcommand{\RNN}{\mathrm{RNN}}

\begin{document}
\title{Lie-Access Neural Turing Machines}
\author{Greg Yang  and  Alexander M. Rush \\
\texttt{\small \{gyang@college,srush@seas\}.harvard.edu}\\
Harvard University\\
Cambridge, MA 02138, USA \\
}

\makeatletter
\let\theauthor\@author
\let\thetitle\@title
\makeatother
\maketitle

\begin{abstract}
  External neural memory structures have recently become a popular tool for
  algorithmic deep learning
  \citep{graves_neural_2014,weston_memory_2014}.  These models
  generally utilize differentiable versions of traditional discrete
  memory-access structures (random access, stacks, tapes) to provide
  the storage necessary for computational tasks.  In
  this work, we argue that these neural memory systems lack specific
  structure important for relative indexing, and propose an
  alternative model, Lie-access memory, that is explicitly designed
  for the neural setting.  In this paradigm, memory is accessed using
  a continuous head in a key-space manifold. The head is moved via Lie
  group actions, such as shifts or rotations, generated by a
  controller, and memory access is performed by linear smoothing in
  key space. We argue that Lie groups provide a natural generalization
  of discrete memory structures, such as Turing machines, as they
  provide inverse and identity operators while maintaining
  differentiability. To experiment with this approach, we implement
  a simplified Lie-access neural Turing machine (LANTM) with
  different Lie groups.  We find that this approach is able to perform
  well on a range of algorithmic tasks.
\end{abstract}
\section{Introduction}
% \linenumbers
% Recurrent neural networks (RNNs) are powerful devices that, unlike conventional neural networks, are able to keep state across time.
% They achieved great results in diverse fields like machine translation \citep{sutskever_sequence_2014, cho_learning_2014, bahdanau_neural_2014}, speech recognition \cite{graves_speech_2013, cho_describing_2015}, image captioning \citep{mao_deep_2014, karpathy_deep_2014, vinyals_show_2014}, and many others.
% However, despite such advances, traditional RNNs still have trouble maintaining memory for long periods of time, presenting an obstacle to attaining human-like general intelligence.
% Recurrent neural networks (RNNs) has gradually replaced many natural language transduction systems in today's technological world due to results such as \cite{sutskever_sequence_2014} and 
% Indeed, its sequential nature is well adapted to the sequential nature of language, and training methods like RMSProp have proven empirically potent.

Recent work on neural Turing machines (NTMs)
\citep{graves_neural_2014,graves2016hybrid} and memory networks
(MemNNs) \citep{weston_memory_2014} has repopularized the use of
explicit external memory in neural networks and demonstrated that
these networks can be effectively  trained in an end-to-end fashion.
These methods have been successfully applied to question answering
\citep{weston_memory_2014, sukhbaatar_end--end_2015, kumar_ask_2015},
algorithm learning \citep{graves_neural_2014, kalchbrenner_grid_2015,
  kaiser_neural_2015, kurach_neural_2015, zaremba_reinforcement_2015,
  grefenstette_learning_2015, joulin_inferring_2015}, machine
translation \citep{kalchbrenner_grid_2015}, and other tasks.  This
methodology has the potential to extend deep networks in a
general-purpose way beyond the limitations of fixed-length encodings
such as standard recurrent neural networks (RNNs).

A shared theme in many of these works (and earlier exploration of
neural memory) is to re-frame traditional memory access paradigms to be
continuous and possibly differentiable to allow for 
backpropagation. In MemNNs, traditional \textit{random-access} memory
is replaced with a ranking approach that finds the most likely memory.
In the work of \citet{grefenstette_learning_2015}, classical
\textit{stack-, queue-}, and \textit{deque-based} memories are replaced by soft-differentiable stack, queue, and deque data-structures. In NTMs, \textit{sequential} local-access memory is
simulated by an explicit tape data structure. 

This work questions the assumption that neural memory should mimic the
structure of traditional discrete memory.  We argue that a neural
memory should provide the following: (A) differentiability for
end-to-end training and (B) robust \textit{relative} indexing (perhaps
in addition to random-access). Surprisingly many neural memory systems
fail one of these conditions, either lacking Criterion B, discussed
below, or employing extensions like REINFORCE to work around lack of
differentiability \citep{zaremba_reinforcement_2015}.

We propose instead a class of memory access techniques based around
Lie groups, i.e. groups with differentiable operations, which provide
a natural structure for neural memory access.  By definition, their
differentiability satisfies the concerns of Criterion A. Additionally
the group axioms provide identity, invertibility, and associativity,
all of which are desirable properties for a relative indexing scheme (Criterion B),
and all of which are satisfied by standard Turing machines. Notably
though, simple group properties like invertibility are not satisfied
by \textit{neural} Turing machines, differentiable neural computers,
or even by simple soft-tape machines.  In short, in our method, we
construct memory systems with keys placed on a manifold, and
where relative access operations are provided by Lie groups.
% Assuming
% that the head of our model is also a point in Euclidean space,
% relative access (Criterion C) can be done by applying a group action
% to ``drag'' the head in a relative direction.  The precise definition
% of this operation is established by a predefined group ${\cal
%   G}$.
% Finally we can satisfy Criterion D by ensuring that ${\cal G}$ is a
% Lie group.Further Background is provided in Section~\ref{} and the model
% is described in more detail in Section~\ref{}.

To experiment with this approach, we implement a neural Turing machine
with an LSTM controller and several versions of 
Lie-access memory, which we call Lie-access neural Turing machines (LANTM).  The details
of these models are exhibited in Section~\ref{LieAccessMemory}.\footnote{
Our implementations are available at \url{https://github.com/harvardnlp/lie-access-memory}}
Our
main experimental results are presented in Section~\ref{Experiments}.
The LANTM model is able to learn non-trivial algorithmic tasks such as
copying and permutating sequences with higher accuracy than more
traditional memory-based approaches, and significantly better than
fixed memory LSTM models. The memory structures and key transformation
learned by the model resemble interesting continuous space
representations of traditional discrete memory data structures.

\section{Background: Recurrent Neural Networks with Memory}
\label{background}

This work focuses particularly on recurrent neural network (RNN)
controllers of abstract neural memories. Formally, an RNN is a
differentiable function $\RNN: {\cal X} \times {\cal H} \to {\cal H}$,
where ${\cal X}$ is an arbitrary input space and ${\cal H}$ is the hidden state space.  On input
$(\p{x}1, \ldots, \p{x}T) \in {\cal X}^T$ and with initial state
$\p{h}0 \in {\cal H}$, the RNN produces states
$\p{h}1, \ldots, \p{h}T$ based on the recurrence,
\begin{align*}
 \p{h}t & := \RNN(\p{x}t, \p{h}{t-1}).
\end{align*}
These states can be used for downstream tasks, for example sequence prediction which produces outputs $(\p{y}1, \ldots, \p{y}T)$ based on an additional transformation and prediction layer $\p{y}{t} = F(\p{h}{t})$ such as a linear-layer followed by a softmax.
RNNs can be trained end-to-end by backpropagation-through-time (BPTT) \citep{werbos_backpropagation_1990}.  
In practice, we use long short-term memory (LSTM) RNNs \citep{hochreiter_long_1997}.
 LSTM's hidden state consists of two variables $(\p{c}t, \p{h}t)$, where $\p{h}t$ is also the output to the external world; we however use the above notation for simplicity.

 An RNN can also serve as the controller for an external memory system
 \citep{graves_neural_2014,grefenstette_learning_2015,zaremba_reinforcement_2015},
 which enables: (1) the entire system to carry state over time from
 both the RNN and the external memory, and (2) the RNN controller to
 collect readings from and compute additional instructions to the
 external memory.  Formally, we extend the recurrence to,
\begin{align*}
\p h t  &:= \RNN( [\p{x}{t}; \p \rho {t-1}], \p h {t-1}), \\
\p \Sigma t, \p \rho t &:= \mathrm{RW}(\p \Sigma {t-1}, \p h t), 
\end{align*}
where $\Sigma$ is the abstract memory state, and $\p \rho t$ is the value read from memory, and $h$ is used as an abstract controller command to a
read/write function $\mathrm{RW}$.  Writing occurs in the mutation of
$\Sigma$ at each time step.  Throughout this work, $\Sigma$ will take
the form of an ordered set $\{ (k_i, v_i, s_i) \}_{i}$ where $k_i \in {\cal K}$ is
an arbitrary key, $v_i \in \R^m$ is a memory value, and $s_i \in \R^{+}$
is a memory strength.

In order for the model to be trainable with backpropagation, the
memory function $\mathrm{RW}$ must also be differentiable. Several
forms of differentiable memory have been proposed in the
literature. We begin by describing two simple forms: (neural)
random-access memory and (neural) tape-based memory. For this section,
we focus on the read step and assume $\Sigma$ is fixed.

\paragraph{Random-Access Memory}
Random-access memory consists of using a now standard
attention-mechanism or MemNN to read a memory (our description follows \citet{DBLP:journals/corr/MillerFDKBW16}).
The controller hidden state is used to output a random-access pointer, $q'(h)$ that determines a weighting of memory vectors via dot products with the corresponding keys. This weighting in turn determines the read values via linear smoothing based on a function $w$, 
\begin{align*}
w_i(q, \Sigma) &:=  \frac{\phantom{\sum_j} s_i \exp{\la q,  k_i\ra}}{\sum_j s_j \exp{ \la q, k_j \ra }} \ \ & \rho := \sum_i w_i(q'(h), \Sigma) v_i.
\end{align*}
The final read memory is based on how ``close'' the read pointer was
to each of the keys, where closeness in key space is determined by $w$.

% The controller also outputs a new key/val pair $(k_{l+1}, M_{l+1})$ and adds it to the memory bank
% $$
% \p \Sigma t := \p \Sigma {t-1} \cup \{(k_{l+1}, M_{l+1})\}.$$

\paragraph{Tape-Based Memory}

Neural memories can also be extended to support relative access by maintaining read state.
Following notation from Turing machines, we call this state the {\it head}, $q$.  
In the simplest case the recurrence now has the form,
$$\Sigma', q', \rho = \mathrm{RW}(\Sigma, q, h),$$
and this can be extended to support multiple heads. 

In the simplest case of soft tape-based memory (a naive version of the
much more complicated neural Turing machine), the keys $k_i$ indicate
one-hot positions along a tape with $k_i = \delta_i$. The head $q$ is a probability
distribution over tape positions.  It determines the read
value by directly specifying the weights.  The controller can only
``shift'' the head by outputting a kernel $K(h)= (K_{-1}, K_0,
K_{+1})$ in the probability simplex $\Delta^2$ and applying
convolution.
\begin{align*}
q'(q, h) & := q * K(h),\qquad\text{i.e.} & & q'_j  = q_{j-1} K_{+1} + q_j K_0 + q_{j+1} K_{-1}
\end{align*}

We can view this as the soft version of a single-step discrete Turing machine where the 
kernel can softly shift the ``head'' of the machine one to the left, one to the right, 
or remain in the same location. The value returned can then be computed with linear smoothing as above,
\begin{align*}
 w_i(q, \Sigma) &:= \frac{\phantom{\sum_j} s_i  \la q, k_i \ra}{\sum_{j} s_j \la q, k_j \ra}   \ \ \ \ \   &\rho := \sum_i w_i(q'(q, h), \Sigma)  v_i. 
\end{align*}

% To write, the controller emits a write value $\p V t$ and write strength $\p s t = s(\p h t) \in [0, 1]$, so that
% \begin{align*}
% \p M t_i &:= (\p q {t-1}_i \p s t) \p V t + (1 - \p q {t-1}_i \p s t) \p M {t-1}_i 
% \end{align*}
% which defines the update $\p \Sigma {t-1} \to \p \Sigma t$.
% In particular, if $\p s t = 0$, then the tape contents stay the same.
% If $\p s t = 1$ and $\p q t$ is concentrated at an index $j$, then $\p M t_j = \p V t$ and the tape stays the same elsewhere.
\section{Lie Groups for Memory}

Let us now take a brief digression and consider the standard
(non-neural) Turing machine (TM) and the movement of its head over a
tape. A TM has a head $q \in \mathbb{Z}$ indicating the position on a
tape.  Between reads, the head can move any number of steps left or
right.  Moving $a+b$ steps and then $c$ steps eventually puts the head
at the same location as moving $a$ steps and then $b+c$ steps ---
i.e. the head movement is \textit{associative}.  In addition, the
machine should be able to reverse a head shift, for example, in a
stack simulation algorithm, going from push to pop --- i.e. each head
movement should also have a corresponding \textit{inverse}.  Finally,
the head should also be allowed to stay put, for example, to read a
single data item and use it for multiple time points, an \textit{identity}.

These movements correspond directly to group actions: the
possible head movements should be associative, and contain inverse and
identity elements. This group  acts on the set of possible head
locations.  In a TM, the set of $\Z$-valued head movement acts
on the set of locations on the $\Z$-indexed infinite tape.  By our
reasoning above, if a Turing machine is to store data contents at
points in a general space ${\cal K}$ (instead of an infinite
$\Z$-indexed tape), then its head movements should form a group and
act on ${\cal K}$ via group actions.

For a neural memory system, we desire the network to be (almost
everywhere) differentiable.  The notion of ``differentiable'' groups is
well-studied in mathematics, where they are known as {\it Lie groups},
and ``differentiable group actions'' are correspondingly called {\it
  Lie group actions}.  In our case, using Lie group actions as
generalized head movements on a general key space (more accurately,
manifolds) would most importantly mean that we can take derivatives of
these movements and perform the usual backpropagation algorithm.

%\footnote{One might also wish that if $v$ is ``close'' to $w$, then $d(v^k x, w^k x)$ is small for all $k$, so that the machine may just output an approximation $w$ to $v$ and follow the ``$w$-trail''. But this is impossible in a Lie group, as any neighborhood of the identity generates the group}

\section{Lie-Access Neural Turing Machines}
\label{LieAccessMemory}
% \begin{figure}
% \centering
% \includegraphics[scale=.4]{graphics/external_mem.png}
% \caption{Abstract diagram of an external memory architecture.
% Note that unlike the other diagrams in this paper (which are unfolded in time), the connections displayed here are recurrent.}
% \label{external_mem}
% \end{figure}

These properties motivate us to propose Lie access as an alternative
formalism to popular neural memory systems, such as probabilistic
tapes, which surprisingly do not satisfy invertibility and often do
not provide an identity.\footnote{The Markov kernel convolutional soft
  head shift mechanism proposed in \citet{graves_neural_2014} and
  sketched in Section \ref{background} does not in general have
  inverses. Indeed, the authors reported problems with the soft head
  losing ``sharpness'' over time, which they dealt with by
  sharpening coefficients. In the followup work,
  \citet{graves2016hybrid} utilize a \textit{temporal memory link
    matrix} for actions. They note, ``the operation $L w$ smoothly
  shifts the focus forwards to the locations written ...  whereas
  $L^\top w$ shifts the focus backwards''~but do not enforce this as a
  true inverse. They also explicitly do not include an identity,
  noting ``Self-links are excluded (the diagonal of the link matrix is
  always 0)''; however, they could ignore the link matrix with an
  interpolation gate, which in effect acts as the identity.} Our
Lie-access memory will consist of a set of points in a manifold
${\cal K}$. We replace the discrete head with a continuous head
$q\in {\cal K}$. The head moves based on a set of Lie group
actions $a \in {\cal A}$ generated by the controller. To read
memories, we will rely on a distance measure in this space,
$d: {\cal K} \times {\cal K} \to \R^{\ge 0}$.\footnote{ This metric
  should satisfy a compatibility relation with the Lie group action.
  When points $x, y \in X$ are simultaneously moved by the same Lie
  group action $v$, their distance should stay the same (One possible
  mathematical formalization is that $X$ should be a Riemannian manifold
  and the Lie group should be a subgroup of $X$'s isometry group.):
  $d(vx, vy) = d(x, y).$ This condition ensures that if the machine
  writes a sequence of data along a ``straight line'' at points
  $x, vx, v^2x, \ldots, v^k x$, then it can read the same sequence by
  emitting a read location $y$ close to $x$ and then follow the
  ``$v$-trail'' $y, vy, v^2y, \ldots, v^k y$. } Together these
properties describe a general class of possible neural memory
architectures.

% The Lie-access memory sketched above can be incorporated 
% as the memory system with an RNN controller, a setup 
% we call a Lie-access neural Turing machine (LANTM). 
% In this section, we describe the details of this model. 

Formally a Lie-access neural Turing machine (LANTM) computes the following 
function,   
\[ \Sigma', q' , q'_{(w)}, \rho  := \mathrm{RW}(\Sigma, q, q_{(w)}, h) \] 
\noindent where $q, q_{(w)}\in {\cal K}$ are resp. read and write
heads, and $\Sigma$ is the memory itself. We implement $\Sigma$, as
above, as a weighted dictionary $\Sigma = \{ (k_i, v_i, s_i) \}_i$.

\subsection{Addressing Procedure}

The LANTM maintains a read head $q$ which at every step is first
updated to $q'$ and then used to read from the memory table.  This
update occurs by selecting a Lie group action from ${\cal A}$ which
then acts smoothly on the key space ${\cal K}$. We parametrize the action transformation, $a : {\cal H} \mapsto {\cal A}$ by the hidden state
to produce the Lie action, $a(h) \in {\cal A}$. In the simplest case,
the head is then updated based on this action (here $\cdot$ denotes
group action):
$ q'  := a(h) \cdot q   $.

% Here we describe how the head $q$ is updated to $q'$ at each time 
% step and how it is used to read a new memory.  We describe the abstraction of the process over a fixed Lie action group 
% ${\cal A}$ acting smoothly on the key space

For instance, consider two possible Lie groups: 

(1) A shift group $\R^2$ acting additively on $\R^2$.  This
  means that ${\cal A} = \R^2 $ so that $a(h) = (\alpha, \beta) $ acts upon a head
  $q = (x, y)$ by,
\[a(h) \cdot q = (\alpha, \beta) + (x, y) = (x + \alpha, y + \beta).\]

(2) A rotation group $SO(3)$ acting on the sphere
$S^2 = \{v \in \R^3: \|v\| = 1\}$. Each rotation can be described by
its axis $\xi$ (a unit vector) and angle $\theta$. An action $(\xi, \theta) \cdot q$
is just the appropriate rotation of the point $q$, and is given by
Rodrigues' rotation formula,
\[a(h) \cdot q = (\xi, \theta) \cdot q = q \cos \theta + (\xi \times q) \sin \theta + \xi \la \xi, q\ra( 1- \cos \theta).\]
Here $\times$ denotes cross product.

% Section \ref{ExampleReps} in the Appendix gives example
% implementations for 

% The controller emits 3 things: a {\it candidate key} $\p{\tilde k} t \in \R^n$, a {\it mixing coefficient}, or {\it gate}, $\p g t \in [0, 1]$ (via the sigmoid function), and an action $\p v t \in G$ that we also call {\it step}.
% The gate $g$ mixes the previous key $\p k {t-1}$ with the candidate key to produce a {\it pre-action key} $\p {\bar k} t$, which is transformed by $\p v t$ to produce the final key $\p k t$: (here $\cdot$ denotes group action)
% \begin{align*}
% \p {\bar k} t &:= \p g t \p {\tilde k} t + (1 - \p g t) \p k {t-1}\\
% \p k t & := \p v t \cdot \p{\bar k} t.
% \end{align*}
% Figure (\ref{address_mech}) summarizes the addressing procedure.

% \begin{figure}[t]
% \centering
% \includegraphics[scale=.6]{graphics/addressing_mech.pdf}
% \caption{addressing mechanism.\label{address_mech}}
% \end{figure}

%  but in a
% future work such a function can be added, for example, by modifying the memory
% strengths.

% But in terms of implementation, the memory vectors are
% stored as rows of a dynamically increasing memory array $\p M t$ of size $\p u
% t \times m$, where $t$ denotes the time step. 
% We write $\p M t(i)$ for the $i$th
% row of this array. 
% The accompanying scalar strengths for each vector are also stored in a dynamic
% array $\p S t$ of size $\p u t \times 1$. 

\subsection{Reading and Writing Memories}

Recall that memories are stored in $\Sigma$, each with a key, $k_i$,
memory vector, $v_i$, and strength, $s_i$, and that memories are read
using linear smoothing over vectors based on a key weighting function
$w$, $\rho := \sum_i w_i(q', \Sigma) v_i$ .  While there are many
possible weighting schemes, we use one based on the distance of each
memory address from the head in key-space assuming a metric $d$ on
${\cal K}$. We consider two different weighting functions (1)
inverse-square and (2) softmax. There first uses the polynomial law
and the second an annealed softmax of the squared distances:

\begin{align*}
w^{(1)}_i(q, \Sigma) := \f{\phantom{\sum_j} s_i d( q, k_i)^{-2}}
					{\sum_j s_j d(q, k_j)^{-2}}  & &  w^{(2)}_i(q, \Sigma, T) := \f{ \phantom{\sum_j} s_i \exp(-d(q, k_i )^{2}/T)} {\sum_j s_j \exp(-d( q ,  k_j)^{2}/T )},\\
\end{align*}

where we use the convention that it takes the limit value when
$q \to k_i$ and $T$ is a {\it temperature} that represents the
certainty of its reading, i.e. higher $T$ creates more uniform $w$.

\begin{figure}
\centering
\includegraphics[scale=.6]{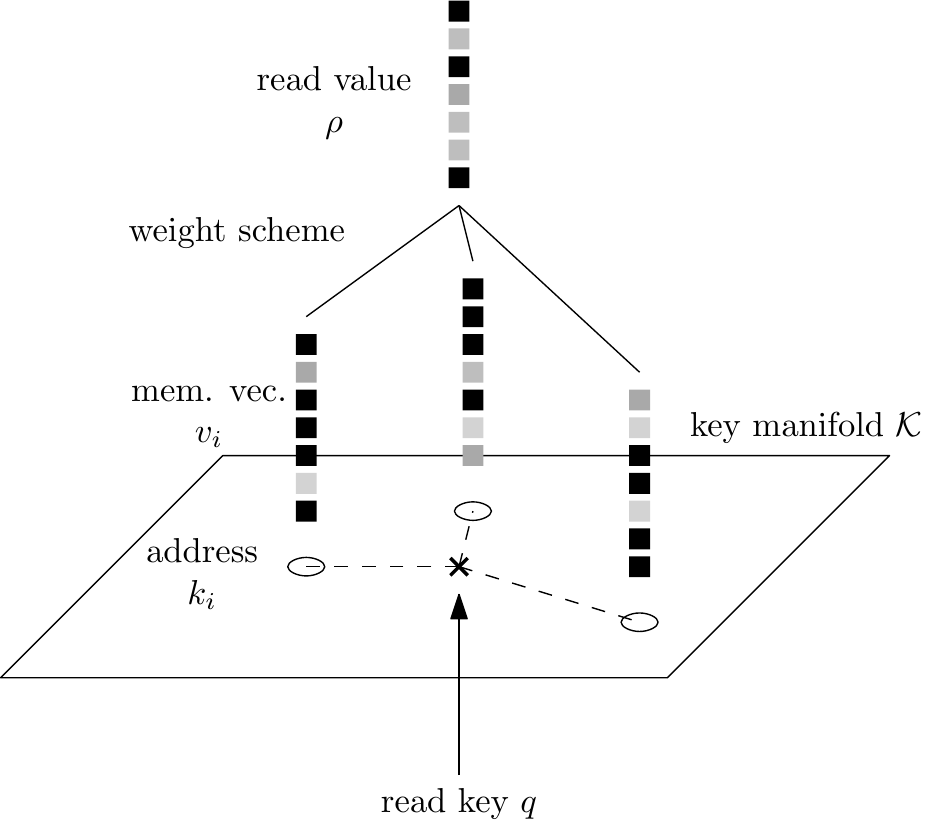}
\caption{\small Retrieval of value from memory via a key. Weightings
  with unit sum are assigned to different memories depending on the
  distances from the addresses to the read key. Linear smoothing
  over values is used to emit the final read value. Both inverse-square and
  softmax schemes follow this method, but differ in their computations of the
  weightings. }
\label{retrieval}
\end{figure}

The writing procedure is similar to reading. The LANTM maintains a
separate \textit{write head} $q_{(w)}$ that moves analogously to the
read head, i.e. with action function $a_{(w)}(h)$ and updated
value $q_{(w)}'$ . At each call to $\mathrm{RW}$, a new memory is
automatically appended to $\Sigma$ with $k = q'_{(w)}$. The corresponding memory
$v$ and strength $s$ are created by MLP's $v(h) \in \R^m$ and $s(h) \in [0, 1]$
taking $h$ as input. After writing, the new memory set is,
$$\Sigma' := \Sigma \cup \{(q'_{(w)}, v(h), s(h))\}.$$

No explicit erase mechanism is provided, but to erase a memory $(k, v, s)$, the controller may in theory write $(k, -v, s)$.
% There are no extra ingredients to
% writing other than adding the produced memory vector $v$, its
% strength $s$, and its address $k$ to the collection of
% memory vectors, strengths, and addresses.  To ensure that memory
% selection by weighted average works well, we squash the values of
% $\p m t$ to $[-1, 1]$ by $\tanh$, but squashing by the logistic
% sigmoid function is also conceivable.  Without such squashing, a
% memory vector $\p M t (i)$ with large values can dominate the output
% of a weight method despite having low weight $\p w t_r (i)$.

\subsection{Combining with Random Access}
\label{interpolations}
% For readers unfamiliar with the Lie group examples mentioned below, we recommend a visit to section \ref{ExampleReps} in the Appendix.

Finally we combine this relative addressing procedure with direct
random-access to give the model the ability for absolute address
access. We do this by outputting an absolute address each step
and simply interpolating with our current head. Write $t(h) \in [0, 1]$ for
the interpolation gate and ${\tilde q}(h) \in {\cal K}$ for our
proposed random-access layer.  For key space manifolds $\mathcal K$
like $\R^n$, \footnote{Or in general, manifolds with convex embeddings
  in $\R^n$.} there's a well defined straight-line interpolation
between two points, so we can set
\begin{align*}
 q' & := a \cdot (t q  + (1- t) {\tilde q})
\end{align*}
where we have omitted the implied dependence on $h$.
For other manifolds like the spheres $S^n$ that have well-behaved projection functions $\pi: \R^n \to S^n$, we can just project the straight-line interpolation to the sphere:
\begin{align*}
 q' & := a \cdot \pi(t q  + (1- t) {\tilde q}).
\end{align*}
In the case of a sphere $S^n$, $\pi$ is just $L_2$-normalization.\footnote{Technically, in the sphere case, $\dom \pi = \R^d - \{0\}$. But in practice one almost never gets 0 from a straight-line interpolation, so computationally this makes little difference.}

\section{Experiments}
\label{Experiments}

We experiment with Lie-access memory on a variety of algorithmic
learning tasks. We are particularly interested in: (a) how Lie-access memory
can be trained, (b) whether it can be effectively utilized
for algorithmic learning, and (c) what internal structures the model
learns compared to systems based directly on soft discrete memory. 
In particular Lie access is not equipped with an explicit stack or tape, 
so it would need to learn continuous patterns that capture these properties. 

% We experiment with two Lie groups: $\R^2$ acting on $\R^2$ by translation and $\SO(3)$ acting on $\S^2$ by rotation, along with the two weight methods InvNorm and SoftMax.
% We outline the most important experimental setup in the main text below but defer other details to the Appendix section \ref{exp-details}. 

% \subsection{Permutation and Arithmetic Tasks}

{\bf Setup.}  Our experiments utilize an LSTM controller in a version
of the encoder-decoder setup \citep{sutskever_sequence_2014}, i.e. an
encoding input pass followed by a decoding output pass. The encoder
reads and writes memories at each step; the decoder only reads
memories.  The encoder is given $\la s\ra$, followed by an the input
sequence, and then $\la /s\ra$ to terminate input. The decoder is not
re-fed its output or the correct symbol, i.e. we do not use teacher
forcing, so $\p{x}t$ is a fixed placeholder input symbol. The decoder
must correctly emit an end-of-output symbol $\la /e \ra$ to terminate.
% Figure (\ref{example_inout}) is an example of inputs and correct outputs during a copy task.

{\bf Models and Baselines.} We implement three main baseline models
including: (a) a standard \textit{LSTM} encoder-decoder, without
explicit external memory, (b) a random access memory network,
\textit{RAM} using the key-value formulation as described in the
background, roughly analogous to an attention-based encoder-decoder,
and (c) an interpolation of a \textit{RAM/Tape}-based memory network
as described in the background, i.e. a highly simplified version of a
true NTM \citep{graves_neural_2014} with a sharpening parameter. Our
models include four versions of Lie-access memory. The main model,
\textit{LANTM}, has an LSTM controller, with a shift group
${\cal A} = \R^2$ acting additively on key space ${\cal K} = \R^2$.
We also consider a model $\textit{SLANTM}$ with spherical memory,
utilizing a rotation group $\mathcal A = SO(3)$ acting on keys in the
sphere $\mathcal K = S^2$. For both of the models, the distance
function $d$ is the Euclidean ($L_2$) distance, and we experiment with
smoothing using \textit{inverse-square} (default) and with an annealed
\textit{softmax}.\footnote{
Note that the read weight calculation of a SLANTM with softmax is essentially the same as the RAM model: For head $q$, $\exp(-d(q, k_i)^2/T) = \exp(-\|q - k_i\|^2/T) = \exp(-(2 - 2 \la q, k_i\ra)/T)$, where the last equality comes from $\|q\| = \|k_i\| = 1$ (key-space is on the sphere). Therefore the weights $w_i = \f{ s_i \exp(-d(q, k_i )^{2}/T)}{\sum_j s_j \exp(-d( q,  k_j)^{2}/T )} =
\f{ s_i \exp(-2\la q, k_i\ra/T)}{\sum_j s_j \exp(-2\la q,  k_j\ra/T )}$, which is the RAM weighting scheme.}

{\bf Model Setup.}
For all tasks, the LSTM baseline has 1 to 4 layers, each with 256 cells.
Each of the other models has a single-layer, 50-cell LSTM controller, with memory width (i.e. the size of each memory vector) 20.
Other parameters such as learning rate, decay, and intialization are found through grid search.
Further hyperparameter details are give in the appendix.

 {\bf Tasks.}  Our experiments are on a series of algorithmic tasks shown
 in Table~\ref{tab:task}. The \textsc{Copy}, \textsc{Reverse}, and \textsc{Bigram Flip}
 tasks are based on \citet{grefenstette_learning_2015}; the \textsc{Double}
 and \textsc{Interleaved Add} tasks are designed in a similar vein. Additionally we
 also include three harder tasks: \textsc{Odd First}, \textsc{Repeat Copy}, and \textsc{Priority Sort}.
 In \textsc{Odd First}, the model must output the odd-indexed elements first, followed by the even-indexed elements.
 In \textsc{Repeat Copy}, each model must repeat a sequence of length 20, $N$ times. In
 \textsc{Priority Sort}, each item of the input sequence is given a priority, and the model must output them in priority order.

We train each model in two regimes, one with a small number of samples (16K) and one with a large number of samples (320K).
In the former case, the samples are iterated through 20 times, while in the latter, the samples are iterated through only once.
Thus in both regimes, the total training times are the same.  Training is done by minimizing negative log likelihood with
 RMSProp.

\begin{table*}[h]

\begin{subfigure}{\textwidth}
\footnotesize
\centering
\begin{tabular}{lllcc}
\toprule
  Task       & Input & Output & Size $k$ & $|{\cal V}|$ \\ \midrule
1 - \textsc{Copy}       & $a_1 a_2 a_3 \cdots a_k$ & $a_1 a_2 a_3 \cdots a_k$& $[2, 64]$ & 128 \\
2 - \textsc{Reverse}    & $a_1 a_2 a_3 \cdots a_k$ & $a_k a_{k-1} a_{k-2} \cdots a_1$& $ [2, 64]$ & 128\\
3 - \textsc{Bigram Flip} & $a_1 a_2 a_3 a_4 \cdots a_{2k-1} a_{2k}$ & $a_2 a_1 a_4 a_3 \cdots a_{2k} a_{2k-1}$& $ [1, 16]$ & 128 \\
4 - \textsc{Double} & $a_1a_2\cdots a_k$ & $2 \times |a_k \cdots a_1|$ & $ [2, 40]$ & 10\\
5 - \textsc{Interleaved Add} &$a_1a_2a_3a_4 \cdots a_{2k-1}a_{2k}$& $|a_{2k}a_{2k-2} \cdots a_2| + |a_{2k-1}\cdots a_1|$ & $[2, 16]$ & 10\\
6 - \textsc{Odd First}   & $a_1 a_2 a_3 a_4 \cdots a_{2k-1} a_{2k}$ & $a_1 a_3 \cdots a_{2k-1} a_2 a_4 \cdots a_{2k}$ & $[1, 16]$ & 128 \\
7 - \textsc{Repeat Copy} & $\overline N a_1 \cdots a_{20}$ & $a_1 \cdots a_{20}  \cdots a_1 \cdots a_{20}$ ($N$ times) & $N \in [1, 5]$ & 128\\
8 - \textsc{Priority Sort} & $\overline 5 a_5 \overline 2 a_2 \overline 9 a_9 \cdots$ & $a_1 a_2 a_3 \cdots a_k $ & $[2, 10]$ & 128\\
\bottomrule      
\end{tabular}
\caption{\label{tab:task}Task descriptions and parameters.
$|a_k\cdots a_1|$ means the decimal number repesented by decimal digits $a_k \cdots a_1$. 
Arithmetic tasks have all numbers formatted with the least significant digits \emph{on the left} and with zero padding.
The \textsc{Double} task takes an integer $x \in [0, 10^k)$ padded to $k$ digits and outputs $2x$ in $k+1$ digits, zero padded to $k+1$ digits. 
The \textsc{Interleaved Add} task takes two integers $x, y \in [0, 10^k)$ padded to $k$ digits and interleaved, forming a length $2k$ input sequence and outputs $x+y$ zero padded to $k+1$ digits.
The last two tasks use numbers in unary format:
$\overline N$ is the shorthand for a length $N$ sequence of a special symbol $@$, encoding $N$ in unary, e.g. $\overline 3 = @@@$.
}
\end{subfigure}

\vspace{0.25cm}
\begin{subfigure}{\textwidth}
\small
\centering
\begin{tabular}{lc@{\,~}@{\,~}c@{\,~}@{\,~}c@{\,~}@{\,~}c@{\,~}@{\,}c@{\,~}@{\,}c@{\,}@{\,}c@{\,}@{\,}c@{\,}@{\,}c@{\,}@{\,}c@{\,}@{\,~}c@{\,~}@{\,~}c@{\,~}@{\,~}c@{\,~}@{\,~}c@{\,~}}
% \cline{2-6}
% &\multicolumn{2}{@{\,}c@{\,}|}{{Weakly}} 
\toprule
 &\multicolumn{2}{c}{{Base}} 
 &\multicolumn{4}{@{\,}c@{\,}}{Memory}
 &\multicolumn{8}{@{\,}c@{\,}}{{Lie}} \\
% \cline{2-6}
 & 
     % {\rotatebox[origin=l]{0}{~~~~~~~~~~~~~~~~~~~~$\task\limits_{\mbox{~~~~~~~~~}}$}} &

\multicolumn{2}{c}{LSTM} &  
    \multicolumn{2}{c}{ RAM} &
     \multicolumn{2}{c}{RAM/Tape} &
     \multicolumn{2}{c}{{LANTM}} &
     \multicolumn{2}{c}{{LANTM-s}} &
     \multicolumn{2}{c}{{SLANTM}} &
     \multicolumn{2}{c}{SLANTM-s} \\
     \midrule
 & S & L & S & L & S & L & S & L & S & L & S & L & S & L\\
\midrule
1 & 16/0  & 21/0  & 61/0  & 61/1  & 70/2  & 70/1  & $\star$  & $\star$  & $\star$  & $\star$  & $\star$  & $\star$  & $\star$  & $\star$ \\
2 & 26/0  & 32/0  & 58/2  & 54/2  & 24/1  & 43/2  & $\star$  & $\star$  & 97/44  & 98/88  & 99/96  & $\star$  & $\star$  & $\star$ \\
3 & 30/0  & 39/0  & 56/5  & 54/9  & 64/8  & 69/9  & $\star$  & $\star$  & $\star$  & 99/94  & 99/99  & 97/67  & 93/60  & 90/43 \\
4 & 44/0  & 47/0  & 72/8  & 74/15  & 70/12  & 71/6  & $\star$  & $\star$  & $\star$  & $\star$  & $\star$  & $\star$  & $\star$  & $\star$ \\
5 & 60/0  & 61/0  & 74/13  & 76/17  & 77/23  & 67/19  & {\bf 99/93}  & 99/93  & 90/38  & 94/57  & 99/91  & 99/97  & 98/78  & $\star$ \\
6 & 29/0  & 42/0  & 31/5  & 46/4  & 43/8  & 62/8  & {\bf 99/91}  & {\bf 99/95}  & 90/29  & 50/0  & 49/7  & 56/8  & 74/15  & 76/16 \\
7 & 24/0  & 37/0  & 98/56  & {\bf 99/98}  & 71/18  & 99/93  & 67/0  & 70/0  & 17/0  & 48/0  & {\bf 99/91}  & 99/78  & 96/41  & 99/51 \\
8 & 46/0  & 53/0  & 60/5  & 80/22  & 78/15  & 66/9  & 87/35  & 98/72  & 99/95  & 99/99  & $\star$  & 99/99  & 98/79  & $\star$ \\
\bottomrule
\end{tabular}
\caption{Main results. Numbers represent the accuracy percentages on
  the fine/coarse evaluations on the out-of-sample $2\times$ tasks.
  The S and L columns resp. indicate small and large sample training
  regimes.  Symbol $\star$ indicates exact $100\%$ accuracy (Fine
  scores above 99.5 are not rounded up). Baselines are described in
  the body.  LANTM and SLANTM use inverse-square while LANTM-s and
  SLANTM-s use softmax weighting scheme.  The best scores, if not
  100\% (denoted by stars), are bolded for each of the small and large
  sample regimes.}

\label{tab:main}

\end{subfigure}

\end{table*}

% \begin{table}[ht]

% \parbox{.4\linewidth}{
% \centering
% \caption{Input/output examples for arithmetic tasks}
% \scriptsize {
% \begin{tabular}{lllll}
% \toprule
% task       & input & output & explanation\\ \midrule
% double       & $928$  & $8561$ & $2 * 829 = 1658$\\
% addition    & $439204$ & $7150$ & $423 + 94 = 517$\\
% \bottomrule      
% \end{tabular}
% }
% \label{arithmetic_inputoutput}
% }
% \parbox{.59\linewidth}{

% \centering
% \caption{Input sequence lengths or operand digits for each task.}
% \scriptsize {
% \begin{tabular}{lllll}
% \toprule
% task       & min train & max train & min test & max test \\ \midrule
% copy       & 2         & 64        & 65       & 128      \\
% reverse    & 2         & 64        & 65       & 128      \\
% bigramFlip & 2         & 32        & 33       & 64       \\
% double     & 2         & 40        & 41       & 80       \\
% addition   & 2         & 16        & 17       & 32		\\
% \bottomrule      
% \end{tabular}
% }
% \label{input_sizes}}
% \end{table}

% \begin{figure}[t]
% \centering
% \includegraphics[scale=.8]{graphics/example_inout_schematics.pdf} 
% \caption{example in/out schematic. \label{example_inout}}
% \end{figure}
% For each task, we train each model with the same number of samples.
% 

Prediction is performed via argmax/greedy prediction at each step. To
evaluate the performance of the models, we compute the fraction of
tokens correctly predicted and the fraction of all answers
completely correctly predicted, respectively called fine and coarse scores.
% Fine score refers to the
% percentage of digit or characters (including the end markers) that the
% model correctly outputs.  Coarse score refers to the percentage of
% total problems that the model answers completely correctly.
We assess the models on 3.2K randomly generated \textit{out-of-sample} 2x
length examples, i.e. with sequence lengths $2k$ (or repeat number $2N$ in
the case of \textsc{Repeat Copy}) to test the generalization of the
system. More precisely, for all tasks other than repeat copy, during training, the length $k$ is varied in the interval $[l_k, u_k]$ (as shown in table \ref{tab:main}a).
During test time, the length $k$ is varied in the range $[u_k+1, 2u_k]$.
For repeat copy, the repetition number $N$ is varied similarly, instead of $k$.

% We report the models with the best coarse scores on 2x generalization in each grid in Table \ref{table:results}.

{\bf Results.} Main results comparing the different memory systems and
read computations on a series of tasks are shown in
Table~\ref{tab:main}. Consistent with previous work the fixed-memory
LSTM system fails consistently when required to generalize to the 2x
samples, unable to solve any 2x problem correctly, and only able to
predict at most $\sim50\%$ of the symbols for all tasks except
interleaved addition, regardless of training regime.  The RAM
(attention-based) and the RAM/tape hybrid are much stronger baselines,
answering more than $50\%$ of the characters correctly for all but the
\textsc{6-Odd First} task.  Perhaps surprisingly, RAM and RAM/tape
learned the \textsc{7-Repeat Copy} task with almost perfect
generalization scores when trained in the large sample regime.  In
general, it does not seem that the simple tape memory confers much
advantage to the RAM model, as the generalization performances of both
models are similar for the most part, which motivates more advanced 
NTM enhancements beyond sharpening.

% With regard to 2x generalization, RAM does slightly better than RAM/tape on copy, reverse, bigram Flip, and double tasks, but RAM/tape does much better than RAM on the rest --- interleaved add, odd first, repeat copy, and priority sort --- which can be seen as more complex tasks than the previous cohort.
% In our experiments we did not see much of an improvement when switching 
% from RAM only to a RAM+soft-tape system. (This is not to indicate that a soft-tape system cannot be effective, but our 
% naive version likely needs more of the optimization that make the NTM effective). 

The last four columns illustrate the performance of the LANTM
models. We found the inverse-square LANTM and SLANTM models to be the most effective,
achieving $>90\%$ generalization accuracy on most tasks, and together they solve all of the tasks here with $>90\%$ coarse score.
In particular, LANTM is able to solve the \textsc{6-Odd First} problem when no other model can correctly solve $20\%$ of the 2x instances;
SLANTM on the other hand is the only Lie access model able to solve the \textsc{7-Repeat Copy} problem. 
% InvNorm seems
% to give improvement here over the more common softmax approach to
% memory access. We also compare results with the sperical SLANTM
% model. A spherical memory system is a bit atypical, but these results
% demonstrate that different Lie groups can easily be substituted in,
% and trained. Interestingly for SLANTM, the softmax achieves better 
% generalization. SLANTM is particularly effective on task 7 (repeat copy), getting the best 
% generalization accuracies. 

The best Lie access model trained with the small sample regime beats
or is competitive with any of the baseline trained under the large
sample regime.  In all tasks other than \textsc{7-Repeat Copy}, the gap in the
coarse score between the best Lie access model in small sample regime
and the best baseline in any sample regime is $\ge70\%$.  However, in
most cases, training under the large sample regime does not improve
much.  For a few tasks, small sample regime actually produces a model
with better generalization than large sample regime.
%, possibly due to variance of these more complex models with respect to initialization.
We observed in these instances, the generalization error curve under a large sample regime reaches an optimum at around 2/3 to 3/4 of training time, and then increases almost monotonically from there.
Thus, the model likely has found an algorithm that works only for the training sizes; in particular, this phenomenon does not seem to be due to lack of training time.

\begin{figure*}
\begin{subfigure}[t]{0.5\textwidth}
	\centering
    \hspace*{-1cm}\includegraphics[width=1.1\linewidth]{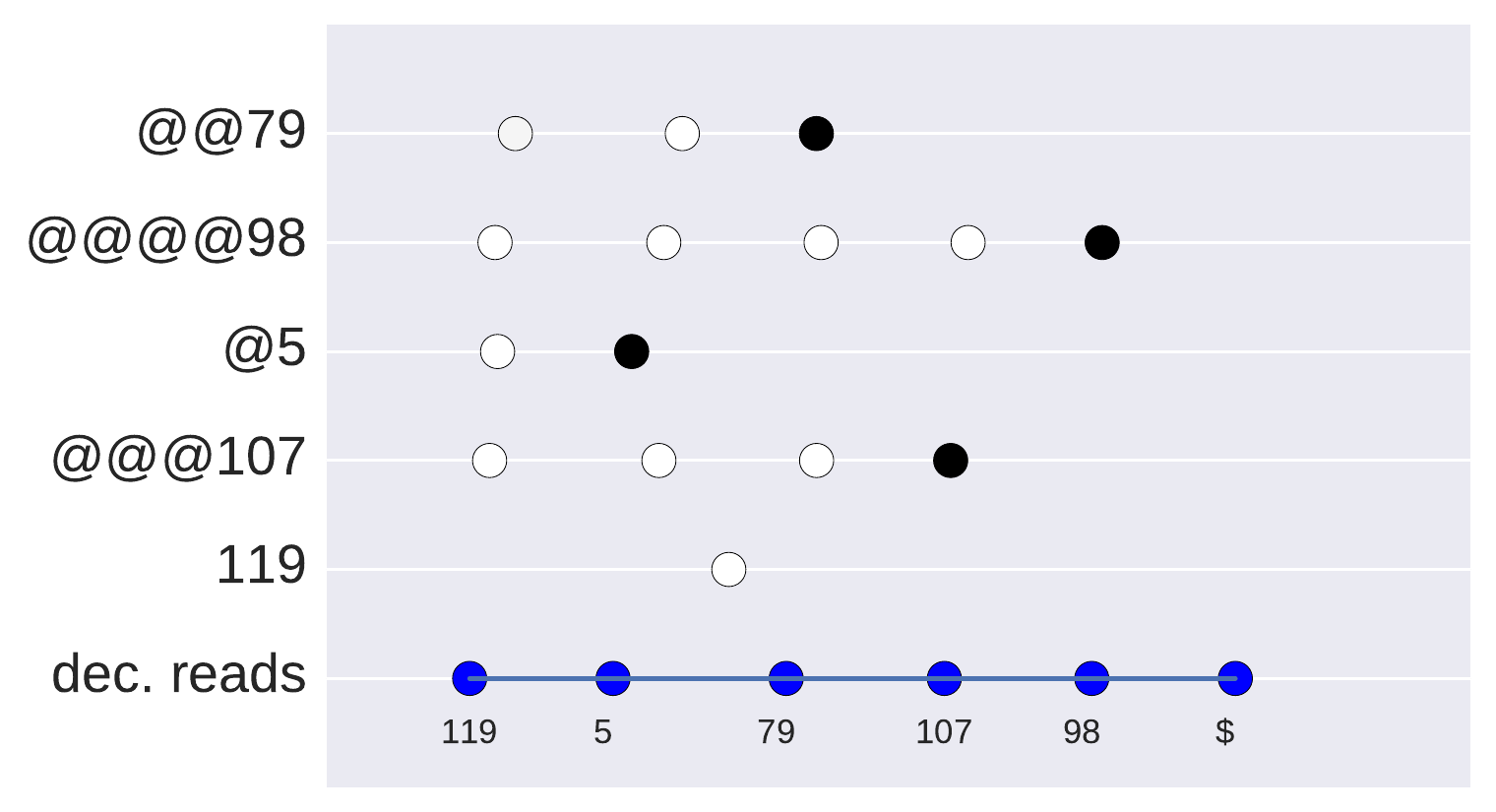}
    \caption{}
\end{subfigure}
    \hspace*{0.2cm}
\begin{subfigure}[t]{0.5\textwidth}
	\centering
    \includegraphics[width=1.1\linewidth]{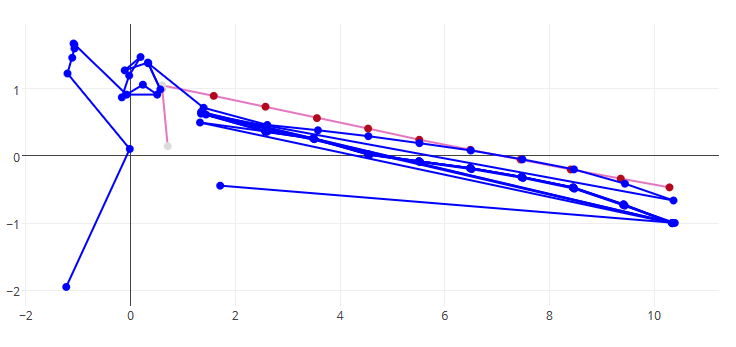}
    \caption{}
\end{subfigure}
  \caption{\label{fig:lantm}\small Analysis of the LANTM model.
    \textbf{(a)} PCA projection from key space $\R^2$ to 1D for the memories $\Sigma$ and read heads $q$ of LANTM for the unary \textsc{8-Priority Sort} task.
    In this task, the encoder reads a priority, encoded in unary, and
    then a value; the decoder must output these values in priority
    order. In this example the sequence is
    $[@,@,79,@,@,@,@,98,@,5,@,@,@,107, @, 119]$, where the special
    symbol @ is a unary encoding of the priority. From top to bottom, each row
    indicates the movement of the encoder write head $q_{(w)}$
    as it is fed each input character.  Fill indicates the strength $s_i$ of memory write (black indicates high strength).
    Position of a dot within its row indicates the PCA projection of the key $k_i$.
    The last line indicates the movement of decoder read head $q$.
    Interestingly, we note that, instead of writing to memory, the controller remembers the item 119 itself.
    \textbf{(b)} Raw coordinates in key space $\R^2$ of writes (red) and reads (blue) from LANTM on \textsc{7-Repeat
    Copy}. Red line indicates the writes, which occur along a straight
    line during the encoding phase. Blue line indicates the reads, which zip
    back and forth in the process of copying the input
    sequence 6 times. }
\end{figure*}
\begin{figure*}
    \centering    
  	\begin{subfigure}[t]{0.48\textwidth}
  	\centering
    \hspace*{-1cm}\includegraphics[width=1.3\linewidth]{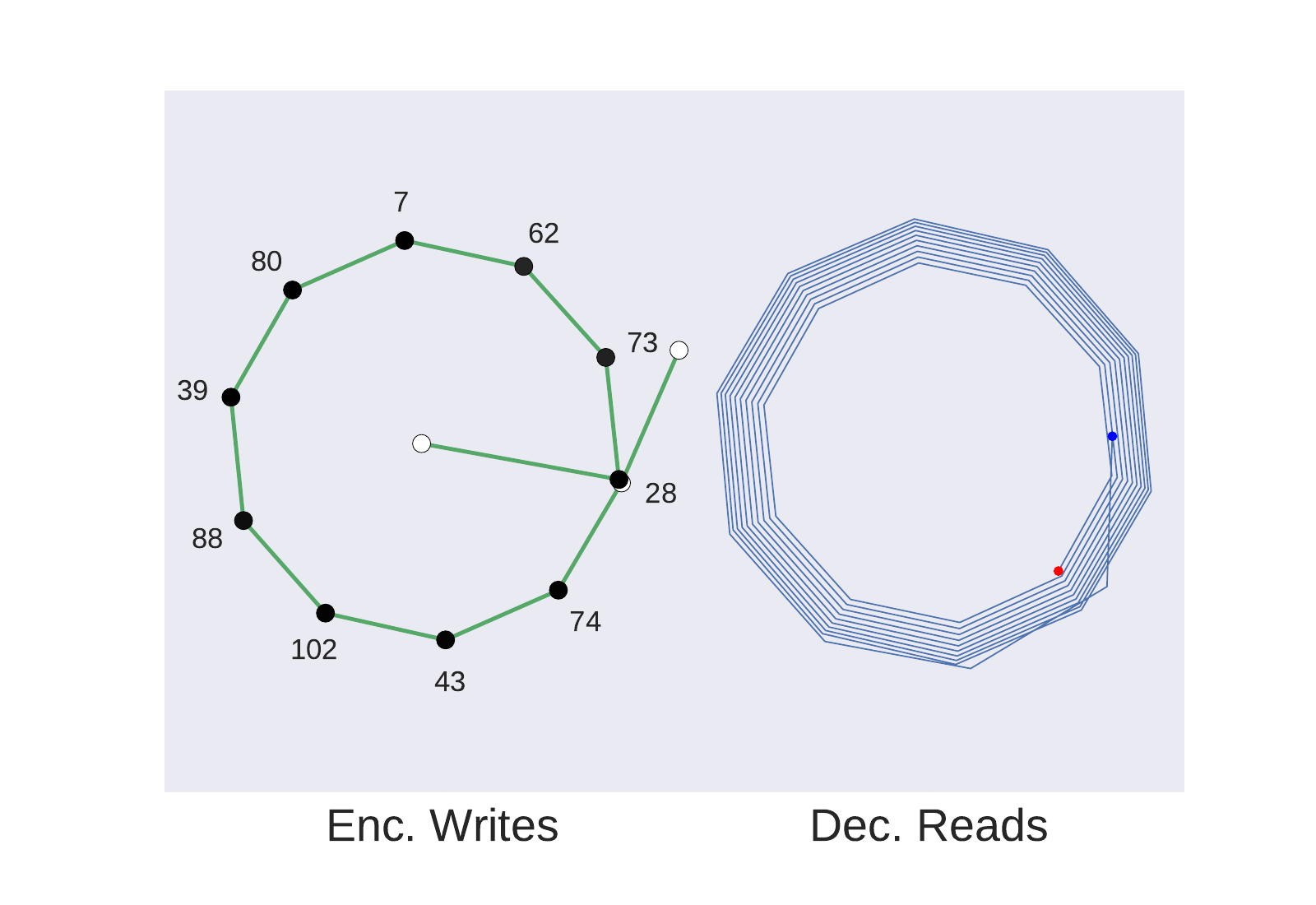}
    \caption{}
    \end{subfigure}
    \begin{subfigure}[t]{0.48\textwidth}
    \includegraphics[width=1.2\linewidth]{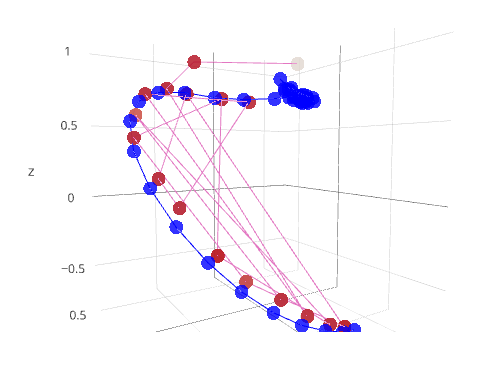}
    \caption{}
    \end{subfigure}
    \caption{\label{fig:slantm}\small Analysis of the SLANTM model. \textbf{(a)} PCA
      projection from the spherical key space $S^2$ to 2D of the memories $\Sigma$ and
      read heads $q$ of SLANTM for the task of \textsc{7-Repeat
        Copy}. Here the model is to repeatedly output the sequence 10
      times. Input is 10 repetitions of special symbol @ followed by
      [28, 74, 43, 102, 88, 39, ... ].
      \textit{Left}: the positions of write head $q_{(w)}$ during the encoding phase.
      Fill indicates strength $s_i$ (black means high strength); number
      indicates the character stored.
      SLANTM traverses in a circle clockwise starting at point 28, and stores data at regular intervals.
      \textit{Right}: the positions of read head $q$ during the
      decoding phase.
      Starting from the blue dot, the reads move
      clockwise around the sphere, and end at the red dot.
	  For the sake of clarity, read positions are indicated by bends in the blue line, instead of by dots.
      Intriguingly, the model implements a cyclic list data structure, taking advantage of the spherical structure of the memory. 
      \textbf{(b)} Raw coordinates in key space $S^2$ of writes (red) and reads (blue) from SLANTM on a non-unary encoded variant of
      the priority sort task. Red line indicates the
      movements of the write-head $q_{(w)}$ to place points along a
      sub-manifold of ${\cal K}$ (an arc of $S^2$) during the encoding phase.
      Notably, this movement is not sequential, but random-access, so as to store elements in correct priority order.
      Blue line indicates the simple traversal of this arc during decoding.}
  \label{fig:trajectory}
\end{figure*}

\begin{figure*}
\centering
\includegraphics[width=.45\textwidth]{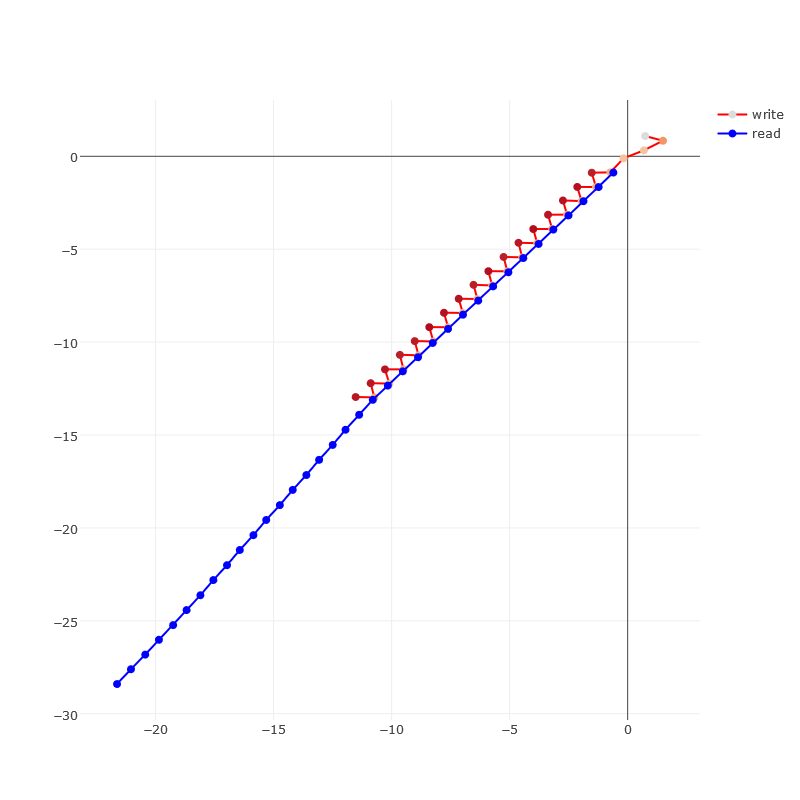}
\includegraphics[width=.45\textwidth]{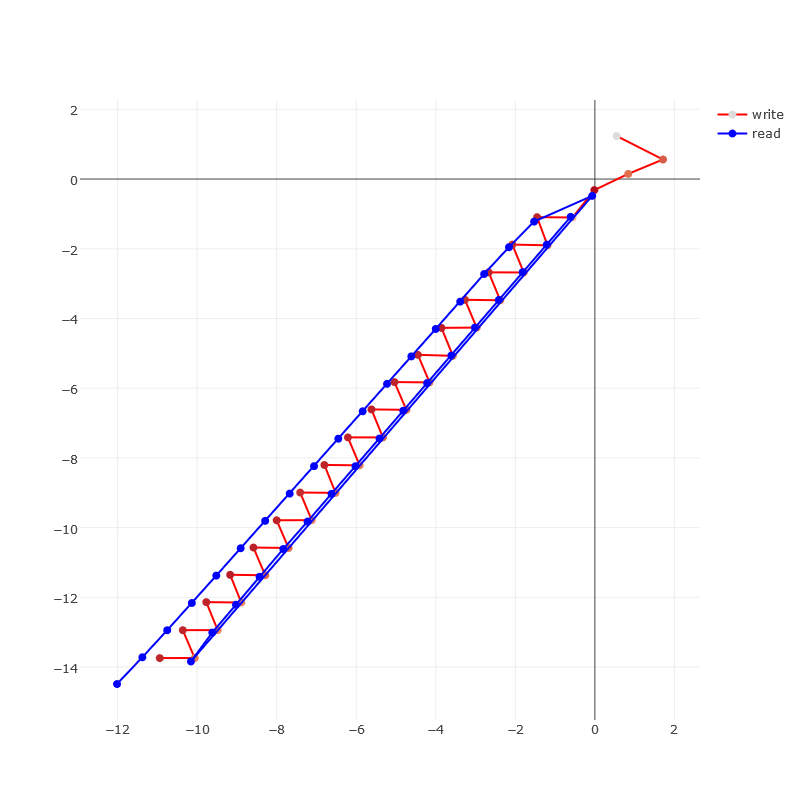}
\caption{\small Memory access pattern of LANTM on \textsc{6-Odd First}. Left:  In the middle of training. LANTM learns to store data in a zigzag such that odd-indexed items fall on one side and even-indexed items fall on the other. However reading is only half correct. Right: After training. During reading, the model simply reads the odd-indexed items in a straight line, followed by the even-indexed items in a parallel line.}
\label{fig:oddfirst}
\end{figure*}

\section{Discussion}

{\bf Qualitative Analysis.} We did further visual analysis of the
different Lie-access techniques to see how the models were learning
the underlying tasks, and to verify that they were using the relative
addressing scheme. Figure~\ref{fig:lantm} shows two diagrams of the
LANTM model of the tasks of priority sort and repeat
copy. Figure~\ref{fig:slantm} shows two diagrams of the SLANTM model
for the same two tasks.
Figure~\ref{fig:oddfirst} shows the memory access pattern of LANTM on \textsc{6-Odd First} task.
Additionally, animations tracing the evolution of the memory access pattern of models over training time can be found at \url{http://nlp.seas.harvard.edu/lantm}. 
They demonstrate that the models indeed learn
relative addressing and internally are constructing geometric data
structures to  solve these algorithmic tasks.

\paragraph{Unbounded storage} One possible criticism of the LANTM
framework could be that the amount of information stored increases
linearly with time, which limits the usefulness of this framework for
long timescale tasks.  This is indeed the case with our implementations,
but need not be the case in general.  There can be many ways of
limiting physical memory usage.  For example, a simple way is to
discard the least recently used memory, as in  the work of \cite{graves2016hybrid}
and \cite{gulcehre_dynamic_2016}.  Another way is to approximate with
fixed number of bits the read function that takes a head position and
returns the read value.  For example, noting that this function is a
rational function on the head position, keys, and memory vectors, we
can approximate the numerators and denominators with a fixed degree
polynomial.

\paragraph{Content address} Our Lie-access framework is not mutually
exclusive from content addressing methods.  For example, in each of
our implementations, we could have the controllers output both a
position in the key space and a content addresser of the same size as
memory vectors, and interpolated the read values from Lie-access and
the read values from content addressing.
 
\section{Conclusion}

This paper introduces Lie-access memory as an alternative neural
memory access paradigm, and explored several different implementations
of this approach.  LANTMs follow similar axioms as discrete Turing
machines while providing differentiability. Experiments show that
simple models can learn algorithmic tasks. Internally these models
naturally learn equivalence of standard data structures like stack and
cyclic lists. In future work we hope to experiment with more groups
and to scale these methods to more difficult reasoning tasks.  For
instance, we hope to build a general purpose encoder-decoder model for
tasks like question answering and machine translation that makes use
of differentiable relative-addressing schemes to replace RAM-style 
attention.

% The LANTM model with the Lie group $\R^2$ acting additively on the key space $\R^2$ and with the InvNorm weight scheme is the strongest model in most of our tasks, but the Lie-access framework is also flexible enough to allow other Lie groups and manifolds, such as the rotation group $SO(3)$ acting on the sphere $S^2$, which produces the best results on the repeat copy and priority sort tasks when equipped with the softmax weight scheme.

\newpage

\bibliography{lantm}
\bibliographystyle{iclr2017_conference}

\newpage
\appendix
\renewcommand\thefigure{\thesection.\arabic{figure}}
\setcounter{table}{0}
\renewcommand*{\thetable}{\thesection.\arabic{table}}

\setcounter{figure}{0}
\appendixpage

\section{Experimental details}
\label{exp-details}

We obtain our results by performing a grid search over the hyperparameters specified in Table \ref{tab:pgrid} and also over seeds 1 to 3, and take the best scores.
We bound the norm of the LANTM head shifts by 1, whereas we try both bounding and not bounding the angle of rotation in our grid for SLANTM.
We initialize the Lie access models to favor Lie access over random access through the interpolation mechanism discussed in section \ref{interpolations}.

The RAM model read mechanism is as discussed in section \ref{background}, and writing is done by appending new $(k, v, s)$ tuples to the memory $\Sigma$.
The only additions to this model in RAM/tape is that left and right keys are now computed using shifted convolution with the read weights:
\begin{align*}
k_L := \sum_i w_{i+1} k_i\\
k_R := \sum_i w_{i-1} k_i
\end{align*}
and these keys $k_L$ and $k_R$ are available (along with the random access key output by the controller) to the controller on the next turn to select from via interpolation.
We also considered weight sharpening in the RAM/Tape model according to \cite{graves_neural_2014}: the controller can output a {\it sharpening coefficient} $\gamma \ge 1$ each turn, so that the final weights are $\tilde w_i = \f{w_i^\gamma}{\sum_j w_j^\gamma}$.
We included this as a feature to grid search over.
\begin{table}[h]
\centering
\small
\begin{tabular}{lllllllll}
\toprule
           & rnn size         & embed & decay delay  & init & learning rate                           & key dim   & custom     \\ \midrule
LANTM(-s)  & $50\times1$              & 14    & \{300, 600\} & \{1, *\} & \{1, 2, 4\}e-2         & 2         & -               \\
SLANTM(-s) & $50\times1$              & 14    & \{300, 600\} & \{1, *\} & \{1, 2, 4\}e-2         & 3         & $\angle$ bound \\
RAM(/tape) & $50\times1$              & 14    & \{300, 600\} & \{1, *\} & \{1, 2, 4\}e-2         & \{2, 20\} & sharpen               \\
LSTM       & $256\times$\{1 to 4\}  & 128   & \{500, 700\} & *        & 2e-\{1 to 4\} & -         & -\\
\bottomrule              
\end{tabular}
\caption{Parameter grid for grid search.
LANTM(-s) means LANTM with invnorm or SoftMax; similarly for SLANTM(-s).
RAM(/tape) means the ram and hybrid ram/tape models.
Initialization: both initialization options set the forget gate of the LSTMs to 1.
The number 1 in the init column means initialization of all other parameters uniformly from $[-1, 1]$.
The symbol * in init column means initialization of all linear layers were done using the torch default, which initializes weights uniformly from $(-\kappa, \kappa)$, where $\kappa$ is $(\text{input size})^{-1/2}$.
For models with memory, this means that the LSTM input to hidden layer is initialized approximately from $[-0.07, 0.07]$ (other than forget gate).
Angle bound is a setting only available in SLANTM.
If angle bound is true, we bound the angle of rotation by a learnable magnitude value.
Sharpening is a setting only available in RAM/tape, and it works as explained in the main text.}
\label{tab:pgrid}
\end{table}

We found that weight sharpening only confers small advantage over vanilla on the \textsc{Copy}, \textsc{Bigram Flip}, and \textsc{Double} tasks, but deteriorates performance on all other tasks.
% Other analysis?

\section{Action Interpolation}

We also experimented with adding an interpolation between the last action
$a^{(t-1)}$ with a candidate action $a(h)$ via a gate
$r(h) \in [0, 1]$ to produce the final action $a^{(t)}$.
Then the final equation of the new head is
$$q' := \p a t \cdot \pi(t q + (1 - t) \tilde q).$$
This allows the controller to easily move in ``a straight
line'' by just saturating both $t$ and $r$.

For example, for the
translation group we have straight-line interpolation, $a^{(t)} := r a + (1 - r) {a^{(t-1)}}$.
For the rotation group $SO(3)$, each rotation is represented by its axis $\xi \in S^{2}$ and angle $\theta \in (-\pi, \pi]$, and we just interpolate each separately
$\xi^{(t)} := \pi(r \xi + (1 - r) {\xi^{(t-1)}})$ and
$\theta^{(t)} := r \theta + ( 1- r) {\theta^{(t-1)}}$.
where $\pi$ is $L_2$-normalization.\footnote{
There is, in fact, a canonical way to interpolate the most common Lie groups, including all of the groups mentioned above, based on the exponential map and the Baker-Campbell-Hausdorff formula \citep{lee_introduction_2012}, but the details are outside the scope of this paper and the computational cost, while acceptable in control theory settings, is too hefty for us.
Interested readers are referred to \cite{shingel_interpolation_2009} and \cite{marthinsen_interpolation_1999}.
}

\begin{figure}
\centering
\includegraphics[width=.4\textwidth]{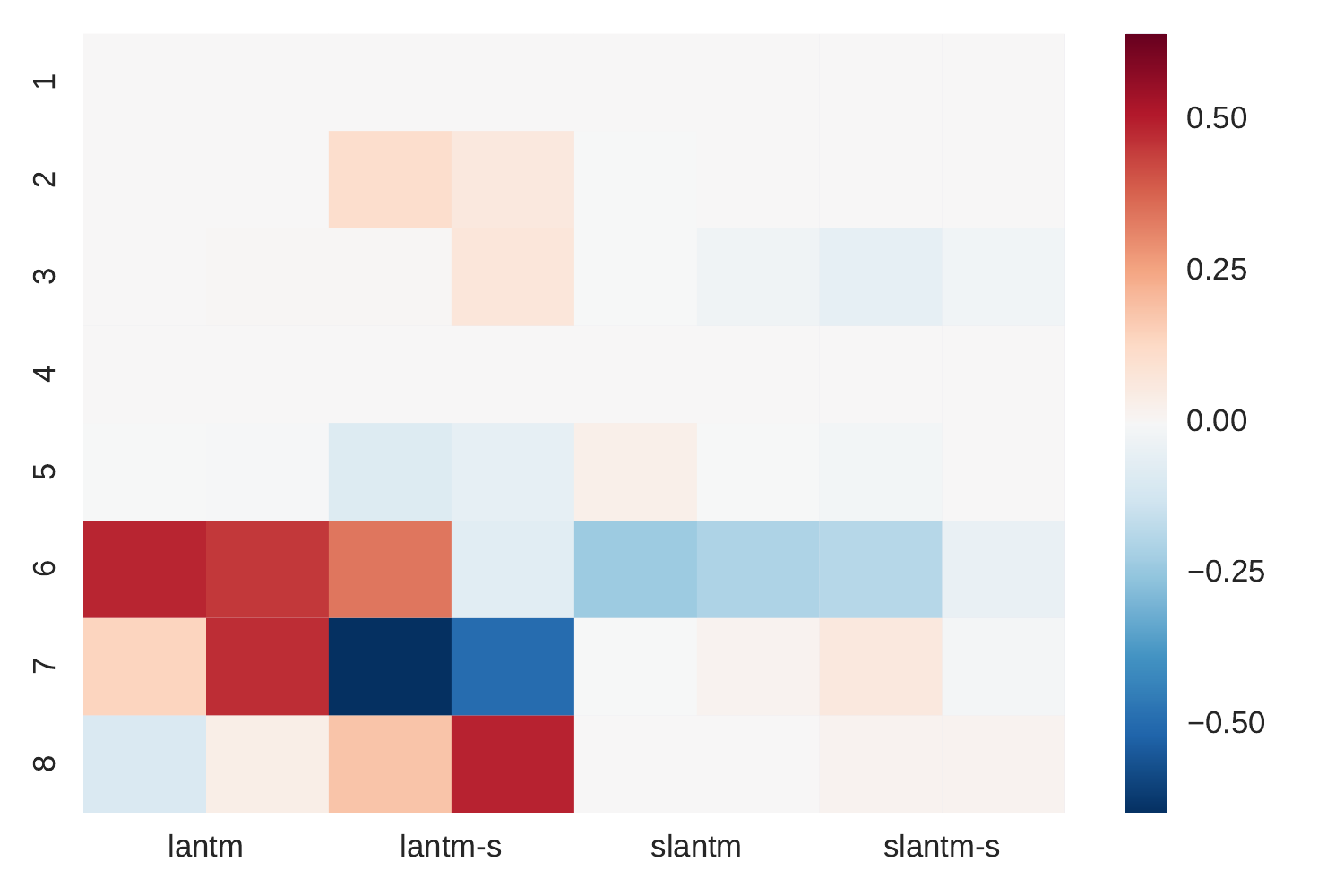}~
\includegraphics[width=.4\textwidth]{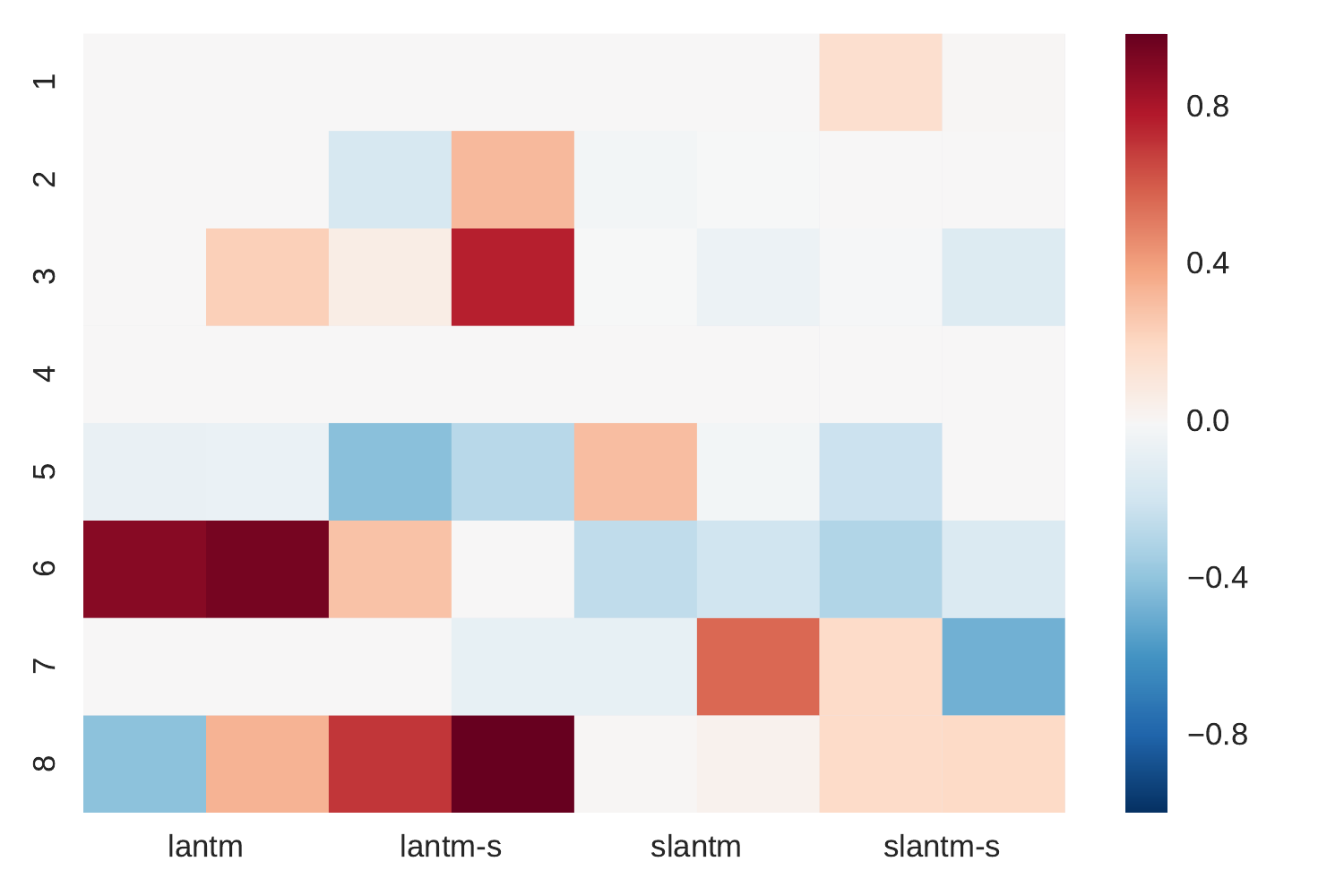}
\caption{The additive difference between the fine (left) and coarse (right) scores of models without action interpolation vs models with action interpolation.
Positive value means model without interpolation performs better.
For each model, the left column displays the difference in small sample regime, while the right column displays the difference in large sample regime.}
\label{fig:diffs}
\end{figure}

We perform the same experiments, with the same grid as specified in the last section, and with the initial action interpolation gates biased toward the previous action.
The results are given in table \ref{tab:act_gate}.
Figure \ref{fig:diffs} shows action interpolation's impact on performance.
Most notably, interpolation seems to improve performance of most models in the \textsc{5-Interleaved Add} task and of the spherical memory models in the \textsc{6-Odd First} task, but causes failure to learn in many situations, most significantly, the failure of LANTM to learn \textsc{6-Odd First}.

\begin{table}
\small
\centering
\begin{tabular}{@{\,}c@{\,}@{\,}c@{\,}@{\,}c@{\,}@{\,}c@{\,}@{\,}c@{\,}@{\,~}c@{\,~}@{\,~}c@{\,~}@{\,~}c@{\,~}@{\,~}c@{\,~}}
% \cline{2-6}
% &\multicolumn{2}{@{\,}c@{\,}|}{{Weakly}} 
\toprule
%  &\multicolumn{2}{c}{{Base}} 
%  &\multicolumn{4}{@{\,}c@{\,}}{Memory}
%  &\multicolumn{8}{@{\,}c@{\,}}{{Lie}} \\
% \cline{2-6}
 & 
     % {\rotatebox[origin=l]{0}{~~~~~~~~~~~~~~~~~~~~$\task\limits_{\mbox{~~~~~~~~~}}$}} &

% \multicolumn{2}{c}{LSTM} &  
%     \multicolumn{2}{c}{ RAM} &
%      \multicolumn{2}{c}{RAM/Tape} &
     \multicolumn{2}{c}{{LANTM}} &
     \multicolumn{2}{c}{{LANTM-s}} &
     \multicolumn{2}{c}{{SLANTM}} &
     \multicolumn{2}{c}{SLANTM-s} \\
     \midrule
 & S & L & S & L & S & L & S & L\\
\midrule \footnotesize
1 & $\star$:$\star$/$\star$:$\star$ & $\star$:$\star$/$\star$:$\star$ & $\star$:$\star$/$\star$:$\star$ & $\star$:$\star$/$\star$:$\star$ & $\star$:$\star$/$\star$:$\star$ & $\star$:$\star$/$\star$:$\star$ & $\star$:99/$\star$:83 & $\star$:99/$\star$:99\\
2 & $\star$:$\star$/$\star$:$\star$ & $\star$:$\star$/$\star$:$\star$ & 97:85/44:60 & 98:91/88:55 & 99:99/96:98 & $\star$:$\star$/$\star$:$\star$ & $\star$:$\star$/$\star$:$\star$ & $\star$:$\star$/$\star$:$\star$\\
3 & $\star$:$\star$/$\star$:$\star$ & $\star$:99/$\star$:77 & $\star$:99/$\star$:93 & 99:92/94:17 & 99:$\star$/99:$\star$ & 97:99/67:73 & 93:99/60:62 & 90:92/43:57\\
4 & $\star$:$\star$/$\star$:$\star$ & $\star$:$\star$/$\star$:$\star$ & $\star$:$\star$/$\star$:$\star$ & $\star$:$\star$/$\star$:$\star$ & $\star$:$\star$/$\star$:$\star$ & $\star$:$\star$/$\star$:$\star$ & $\star$:$\star$/$\star$:$\star$ & $\star$:$\star$/$\star$:$\star$\\
5 & 99:$\star$/93:$\star$ & 99:$\star$/93:$\star$ & 90:99/38:80 & 94:99/57:84 & 99:96/91:61 & 99:99/97:99 & 98:99/78:99 & $\star$:$\star$/$\star$:$\star$\\
6 & 99:50/91:0 & 99:54/95:0 & 90:56/29:0 & 50:57/0:0 & 49:73/7:33 & 56:76/8:27 & 74:92/15:45 & 76:81/16:31\\
7 & 67:52/0:0 & 70:22/0:0 & 17:82/0:0 & 48:98/0:8 & 99:$\star$/91:$\star$ & 99:97/78:21 & 96:90/41:22 & 99:99/51:99\\
8 & 87:97/35:76 & 98:93/72:38 & 99:81/95:24 & 99:50/99:0 & $\star$:99/$\star$:99 & 99:99/99:95 & 98:95/79:60 & $\star$:98/$\star$:80\\

\bottomrule
\end{tabular}
\caption{Comparison between scores of model with action interpolation and without action interpolation.
Numbers represent the accuracy percentages on the fine/coarse evaluations on the out-of-sample $2\times$ tasks.
The S and L columns resp. indicate small and large sample training regimes.  Symbol $\star$ indicates exact $100\%$ accuracy (Fine scores above 99.5 are not rounded up).
Each entry is of the format A:B/C:D, where A and C are respectively the fine and coarse scores of the model without action interpolation (same as in table \ref{tab:main}), and B and C are those for the model with action interpolation. }

\label{tab:act_gate}

\end{table}

\end{document}